\documentclass{article}

\usepackage{microtype}
\usepackage{graphicx}
\usepackage{subfigure}
\usepackage{booktabs} 

\usepackage{hyperref}

\usepackage[accepted]{icml2021}

\icmltitlerunning{Controlling Weather Field Synthesis Using Variational Autoencoders}

\begin{document}

\twocolumn[
\icmltitle{Controlling Weather Field Synthesis Using Variational Autoencoders}




\begin{icmlauthorlist}
\icmlauthor{Dario Augusto Borges Oliveira}{to}
\icmlauthor{Jorge Guevara Diaz}{to}
\icmlauthor{Bianca Zadrozny}{to}
\icmlauthor{Campbell D. Watson}{goo}
\end{icmlauthorlist}

\icmlaffiliation{to}{IBM-Research, Rua Tutoia 1157\\Vila Mariana, SP, Brazil}
\icmlaffiliation{goo}{IBM-Research, 1101 Kitchawan Rd, Yorktown Heights, NY, USA}

\icmlcorrespondingauthor{Dario Augusto Borges Oliveira}{darioaugusto@gmail.com}

\icmlkeywords{Machine Learning, ICML}

\vskip 0.3in
]



\printAffiliationsAndNotice{\icmlEqualContribution} 

\begin{abstract}
One of the consequences of climate change is an observed increase in the frequency of extreme climate events. That poses a challenge for weather forecast and generation algorithms, which learn from historical data but should embed an often uncertain bias to create correct scenarios. This paper investigates how mapping climate data to a known distribution using variational autoencoders might help explore such biases and control the synthesis of weather fields towards more extreme climate scenarios. We experimented using a monsoon-affected precipitation dataset from southwest India, which should give a roughly stable pattern of rainy days and ease our investigation. We report compelling results showing that mapping complex weather data to a known distribution implements an efficient control for weather field synthesis towards more (or less) extreme scenarios.
\end{abstract}

\section{Introduction}
\label{introduction}

As the climate system warms, the frequency, duration, and intensity of different types of extreme weather events have been increasing. For example, climate change leads to more evaporation that may exacerbate droughts and increase the frequency of heavy rainfall and snowfall events \cite{NAS2016}. That directly impacts various sectors such as agriculture, water management, energy, and logistics, which traditionally rely on seasonal forecasts of climate conditions for planning their operations. 

In this context, stochastic weather generators are often used to provide a set of plausible climatic scenarios, which are then fed into impact models for resilience planning and risk mitigation. A variety of weather generation techniques have been developed over the last decades. However, they are often unable to generate realistic extreme weather scenarios, including severe rainfall, wind storms, and droughts \cite{Verdin2018}.

Recently different works proposed to explore deep generative models in the context of weather generation, and most explored generative adversarial networks (GAN) \cite{goodfellow}. \cite{ibm_wg} proposed to use generative adversarial networks to learn single-site precipitation patterns from different locations. \cite{exgan} proposed a GAN-based approach to generate realistic extreme precipitation samples using extreme value theory for modeling the extreme tails of distributions. \cite{precipgan} presented an approach to reconstruct the missing information in passive microwave precipitation data with conditional information. \cite{klemmer2021generative} proposed a GAN-based approach for generating spatio-temporal weather patterns conditioned on detected extreme events.

While GANs are very popular for synthesis in different applications, they do not explicitly learn the training data distribution and therefore depend on auxiliary variables for conditioning and controlling the synthesis. Variational Autoencoders (VAEs) \cite{vaes} are an encoder-decoder generative model alternative that explicitly learns the training set distribution and enables stochastic synthesis by regularizing the latent space to a known distribution. Even if one can also trivially control VAEs synthesis using conditioning variables, such models also enable synthesis control from merely inspecting the latent space distribution to map where to sample to achieve synthesis with known characteristics.

In this paper, we explore VAEs for controlling weather field data synthesis towards more extreme scenarios. We propose to train a VAE model using a normal distribution for the latent space regularization. Then, assuming that extreme events in historical data are also rare, we control the synthesis towards more extreme events by sampling from normal distribution tails, which should hold less common data samples. We report compelling results, showing that controlling the sampling space from a normal distribution implements an effective tool for controlling weather field data synthesis towards more extreme weather scenarios.

\section{Method}

\begin{figure}[h!]
	\centering
		\includegraphics[width=0.8\linewidth]{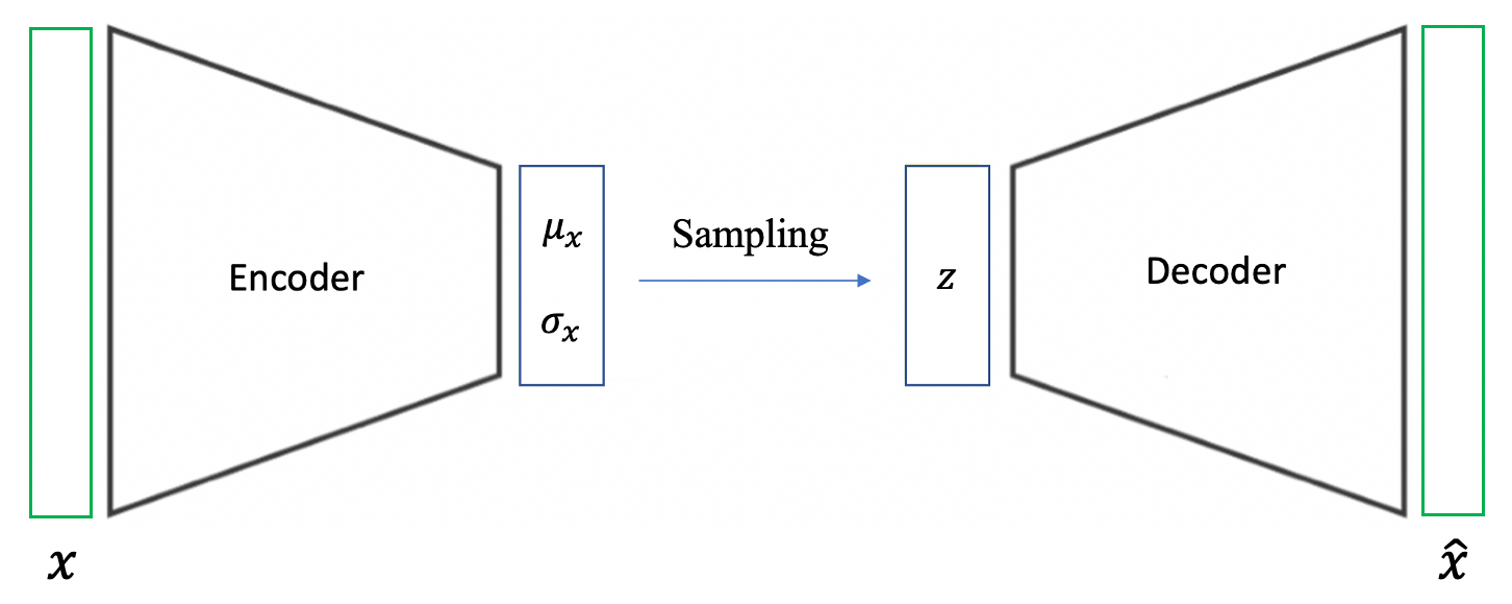}
	\caption{Variational Autoencoders schema. The network is composed by an encoder that maps input data into dense layers representing means ($\mu_x$) and standard deviations ($\sigma_x$), a sampling layer that samples from that distribution and a decoder that maps latent data $z$ into the output.}
	\label{fig:vae}
\end{figure}

VAEs consist of an encoder network that parameterizes a posterior distribution $q(z|x)$ of discrete latent random variables $z$ given the input data $x$, a prior distribution $p(z)$, and a decoder with a distribution $p(x|z)$ over input data, as observed in Figure \ref{fig:vae}. VAEs usually involve two targets, high reconstruction quality and good regularization of the latent space distribution. A common trick for training VAEs is to assume that posteriors and priors in VAEs are normally distributed, which allows simple Gaussian reparametrization for end-to-end training \cite{pmlr-v32-rezende14,vaes}.

Training VAEs usually involves a loss function that combines a reconstruction term that targets building outputs very similar to the inputs, and a regularization term targets creating a latent space with known distribution. The standard loss for a standard VAE \cite{vaes} can be formally described as: 
\noindent
\begin{equation}
    \mathcal{L} = L_{rec} + L_{reg} = \sum_{x\in\mathcal{X}} ||x - \hat{x}|| + \sum_{k}KL(p_k(z|x),N(0,1))
\end{equation}
\noindent
where $x$ is a sample and its corresponding reconstruction is $\hat{x}$;  $KL(p,N(0,1))$ is the KL regularization loss between the latent space distribution $p$ and the standard normal distribution for each $k$ latent space dimension, as described in \cite{vaes}; and $L_{rec}$ is commonly the mean squared error between a sample $x$ and its corresponding reconstruction $\hat{x}$.

\begin{figure}[h!]
	\centering
		\includegraphics[width=1\linewidth]{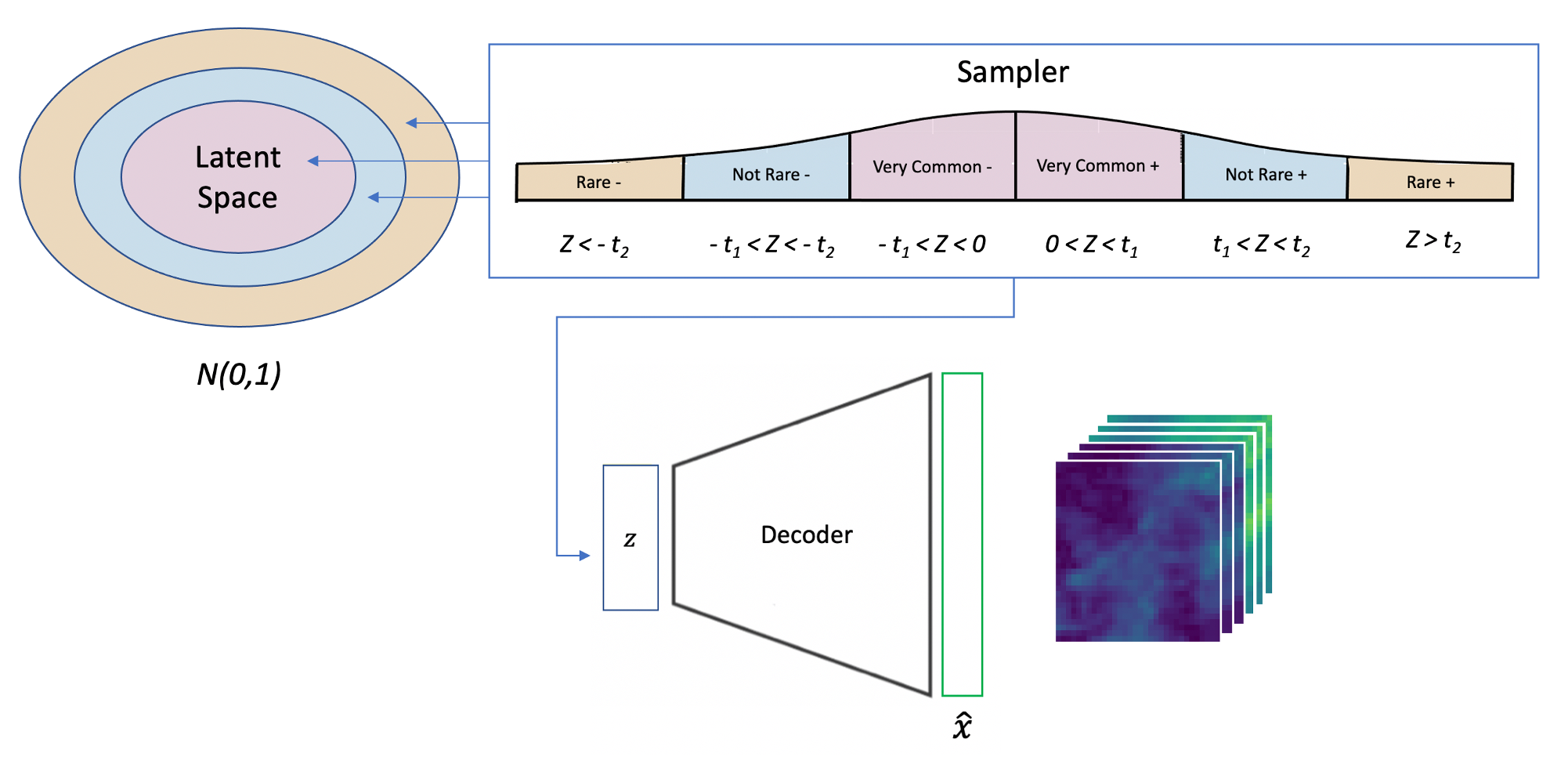}
	\caption{The proposed sampler schema using the standard deviation to define loci in the normal distribution for selecting samples with different characteristics.}
	\label{fig:sampler}
\end{figure}

With a trained decoder model that receives normal-distribution data and decodes into weather field data, we benefit from an inherent property of variational autoencoders training that cluster similar samples close together and assume that regular weather samples will be allocated to the distribution bulk, while less common (including extreme) weather events will be allocated to the distribution tails. That configuration enables us to control the synthesis by simply defining suitable loci in the distribution for sampling. With a rule that the more extreme an event is, the less likely, we come up with a simple synthesis control schema, depicted in Figure \ref{fig:sampler}. We select thresholds $t_i$ that define loci of the normal distribution to sample that are directly related to how the normal distribution probability. The higher $t_i$, the less probable the synthesized sample would be, and supposedly, the more extreme. That simple procedure enables using the latent space mapping to control synthesis and create data coherent with more extreme climate scenarios. 

\section{Experiments}

Our experiments intend to explore how mapping complex weather data with intricate spatio-temporal patterns into a more straightforward latent representation with a known distribution can be efficiently used for controllable extreme scenario weather data synthesis.

\begin{figure}[h!]
	\centering
		\includegraphics[width=1\linewidth]{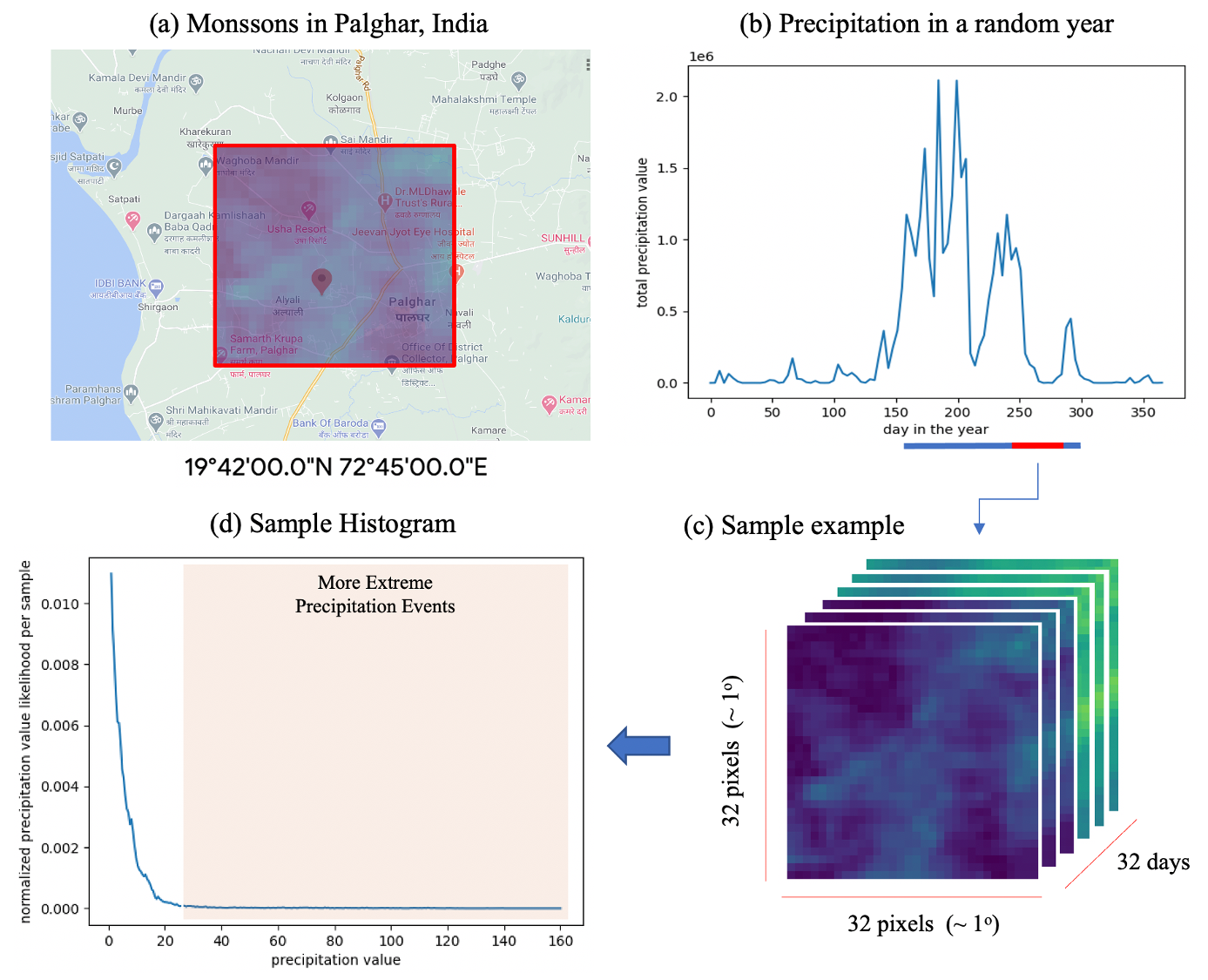}
	\caption{The dataset contains precipitation fields in Palghar, India which includes an annual Monsoon period. (b) represents the distribution of precipitation during a selected year in Palghar with the Monsoon observed between day 150 and day 300. A typical training sample is depicted in (c) and is sampled at random from the Monsoon period and consists of a sequence of 32 days of precipitation data in a squared 1x1 degree tile, which comprises 32 pixels after resizing. In (d) we observe a typical sample histogram.}
	\label{fig:data}
\end{figure}

\subsection{Dataset}

We used the Climate Hazards group Infrared Precipitation with Stations v2.0 (CHIRPS) dataset \cite{funk2015climate}, a global interpolated dataset of daily precipitation providing a spatial resolution of 0.05 degrees. The data ranges from the period 1981 to the present. We experimented with a one-degree by one-degree bounding boxes around the spatial region characterized by the latitude and longitude coordinates $(20,75)$ and $(21,76)$ that geographically corresponds to Palghar, India (as indicated in Figure \ref{fig:data}(a)). We used daily precipitation data from 01/01/1981 to 31/12/2009 as the training data set and the data from 01/01/2010 to 31/12/2019 as the test set.

One can also visually inspect in Figure \ref{fig:data} the histogram for precipitation values in a year and our sampling schema for creating the training set. As indicated in Figure \ref{fig:data}(b), India's monsoon period begins around day 150 in the year and goes on to around day 300. We considered sequences of 32 days in this time range and 16 bounding boxes randomly picked around the coordinate center, and selected 18.000 random samples for composing our final training (14.400) and testing (3.600) sets, as shown in Figure \ref{fig:data}(c). 

\subsection{Experimental Design}

Table \ref{tab:architecture} presents the architecture of the encoder and decoder networks. The encoder architecture consists of two convolution blocks followed by a bottleneck dense layer and two dense layers for optimizing $\mu_x$ and $\sigma_x$ that hold the latent space distribution that is sampled to derive $z$ following a standard normal distribution. We employed two down-sampling stages (one for each convolutional block) to reduce the spatial input dimension by four before submitting the outcome to the bottleneck dense layer. After the convolutional and dense layers, we also applied ReLU activation functions. The decoder receives an input array $z$ with the size of the latent space dimension that is ingested to a dense layer to be reshaped into 256 activation maps of size $8x8x8$. These maps serve as input to consecutive transposed convolution layers that up-sampling the data up to the original size. A final convolution using one filter is applied to deliver the final outcome. 

\begin{figure}[ht!]
	\centering
		\includegraphics[width=1\linewidth]{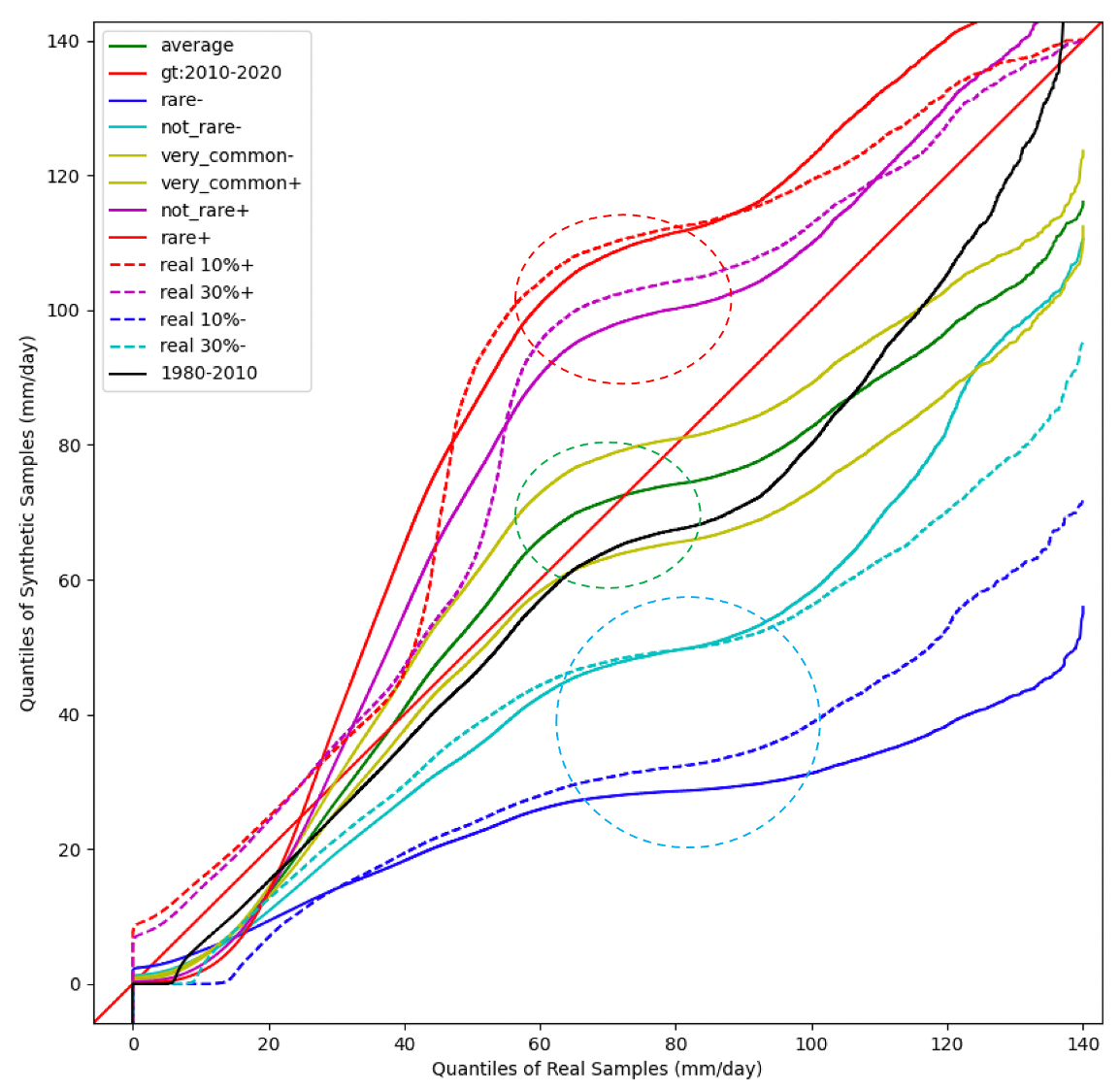}
	\caption{Quantile-quantile plots for training and testing sets. Blue lines represent data with relative lower precipitation values compared to the testing data, red lines higher precipitation scenarios. In black, the historical data as a reference, and in green and yellow scenarios with average precipitation values.}
	\label{fig:qqplot}
\end{figure}

We used the Adam optimizer with a learning rate of $0.001$, beta1 as 0.9, and beta2 as 0.999. We implemented a warm-up period of 10 epochs before considering the regularization term in the loss, which was weighted using the $\beta$-VAE criteria \cite{betavae}. We trained the models for 100 epochs, with 32 data samples per batch, and monitored the total loss to apply early stop. All experiments were carried out using  V100 GPUs. 

\begin{table}[h!]
	\centering
	\caption{Architecture of the networks. Down-sampling is performed in conv 1-1 and conv 1-2 and up-sampling is performed in convt 2-1 and convt 2-2, with a stride of 2.}
	\vspace{5pt}
	\scalebox{0.9}{
	\begin{tabular}{cc|cc}
	    \toprule
		\multicolumn{2}{c|}{\textbf{Encoder}} & \multicolumn{2}{c}{\textbf{ Decoder}} \\
		\multicolumn{2}{c|}{Input $x$ (dim=$32\times32\times32$) } & \multicolumn{2}{c}{Input $z$ (dim=$30$)} \\\midrule
		Layer & Processing & Layer & Processing \\\midrule
		conv 1-1 &$3\times3\times3$, 128 & dense & 131072 \\
		conv 1-2 &$3\times3\times3$, 128 & reshape & $8\times8\times8\times256$ \\
		dense-bn & 500 & convt 2-1 & $3\times3\times3$, 128 \\
		dense-$\mu_x$ & 30 & convt 2-2 & $3\times3\times3$, 128 \\
		dense-$\sigma_x$ & 30 & convt 2-3 & $3\times3\times3$, 1 \\
		sampling & - & - \\
		\bottomrule
    \end{tabular}
    }
    \label{tab:architecture}
\end{table}

\section{Results}

\begin{figure*}[h!]
	\centering
		\includegraphics[width=0.78\linewidth]{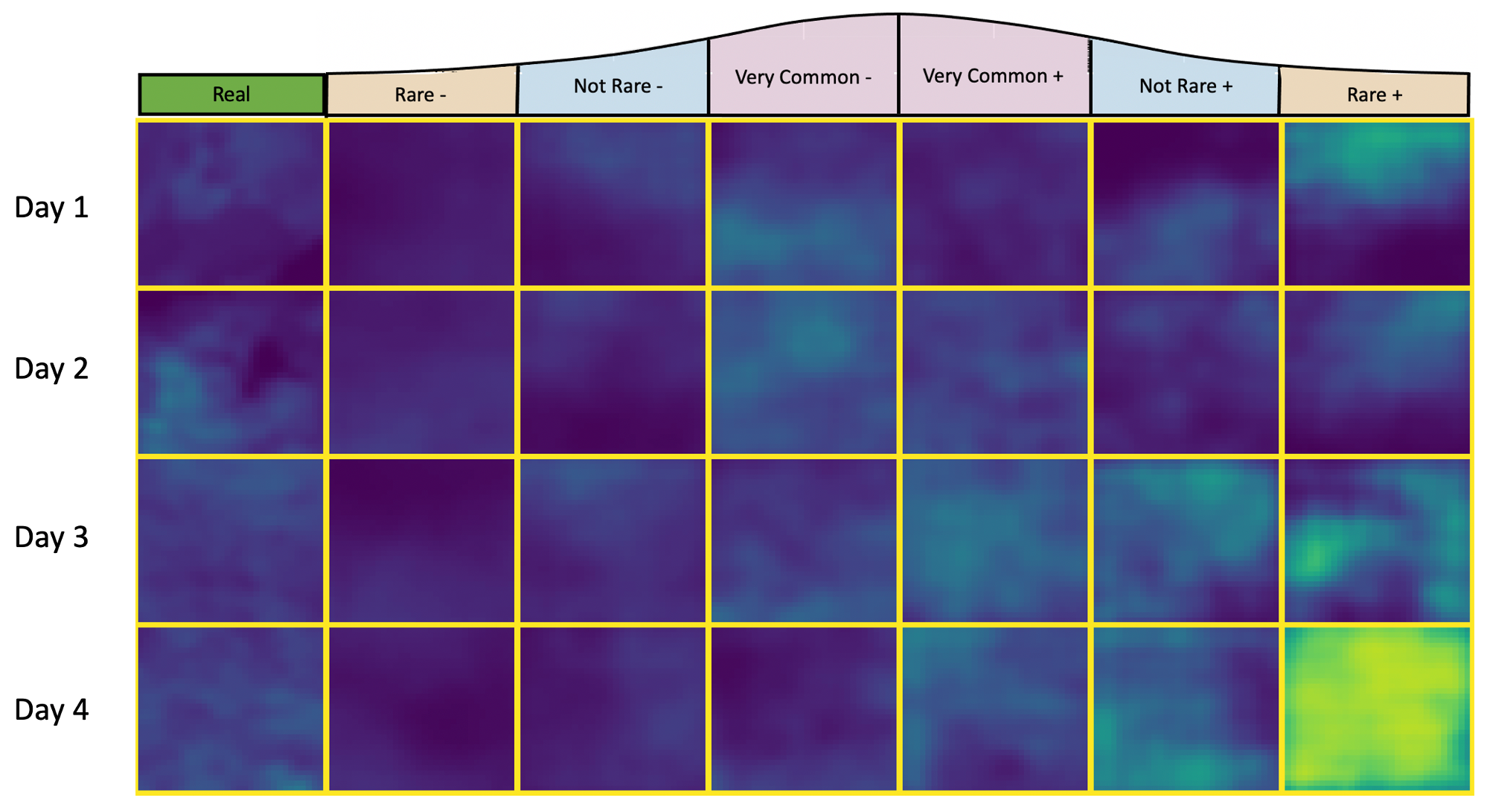}
	\caption{Examples of synthesized samples considering different standard deviation scenarios, and real samples from the testing set as reference. Rows represent four different weather fields selected at random.}
	\label{fig:samples}
\end{figure*} 

For evaluating our results, we used a quantile-quantile (QQ) plot, which is a probability plot used for comparing two probability distributions. In QQ plots, the quantiles of the distributions are plotted against each other, and therefore a point on the plot corresponds to one of the quantiles of a given distribution plotted against the same quantile of another distribution. In our cause, one distribution is computed from input samples pixel values, and the other from the reconstructed samples pixel values. If these two distributions are similar, the plot points will approximately lie on the line where axis-x is equal to axis-y (represented by the straight red line in Figure \ref{fig:qqplot}. If the distributions are linearly related, the points will approximately lie on a line, but not necessarily on the line where axis-x is equal to axis-y. We also plot the historical data distribution to be used as a reference in the test set plots. We also provide synthesized samples for enabling a visual inspection of our results.

Figure \ref{fig:qqplot} compiles a few important information. First, we observe in the black line that the historical data is not precisely defining the test data distribution. That indicates the distribution of precipitation values suffered a shift from the training time interval (1980 to 2010) to the testing (2010 to 2020), especially for higher quantiles, meaning that higher precipitation values became more common than in historical data. One can also observe that our trained vanilla VAE model was able to synthesize data that match the testing distribution up to around 70mm/day but then failed to match the quantiles for higher precipitation values (considering we trained the model using historical data, that is somewhat expected). 

Concerning synthesis control, we created reference extreme weather data, considering the 10\% and 30\% samples with the greatest and lowest average precipitation values, and then evaluated if our sampling schema can control the synthesis towards those samples by simply varying the threshold $t_i$. By quickly varying the standard deviation values from the normal $N(0,\sigma_i$ with $\sigma_i=\{0.3,0.5,0.65,0.75,0.85,1.0,1.3\}$, we were able to synthesize samples which have distributions coherent with those selected as references for more or less extreme weather field data. One can visually inspect in Figure \ref{fig:qqplot} some highlights. The light blue dashed circle depicts scenarios with lower precipitation values, where the dashed line is the curve from selected reference samples, and the continuous lines are those from the synthesized samples using lower standard deviation values. The green dashed circle shows average rain scenarios, and the red dashed circle shows heavier precipitation plots, where the decoded samples using larger standard deviation are coherent with the reference samples for heavier precipitation.

Finally, we also present synthesized samples for visual inspection. In Figure \ref{fig:samples}, samples are selected at random, and it is possible to observe that samples from the average standard deviation sampling are similar to those drawn from real data, as expected since they are more likely to happen. The samples synthesized using smaller standard deviation values depict weather fields with lower precipitation values, and the ones using larger standard deviation seem to show higher precipitation patterns.  

\section{Conclusions}

This paper explored the efficient use of variational autoencoders as a tool for controlling the synthesis of weather fields considering more extreme scenarios. An essential aspect of weather generators is controlling the synthesis for different weather scenarios in light of climate change. We reported that controlling the sampling from the known latent distribution is effectively related to synthesizing samples with more extreme scenarios in the precipitation dataset experimented in our tests. As further research, we expect to explore models that enable multiple distributions for finer control of synthesis and to tackle data with multiple weather system distributions.

\bibliography{example_paper}

\begin{thebibliography}{11}
\providecommand{\natexlab}[1]{#1}
\providecommand{\url}[1]{\texttt{#1}}
\expandafter\ifx\csname urlstyle\endcsname\relax
  \providecommand{\doi}[1]{doi: #1}\else
  \providecommand{\doi}{doi: \begingroup \urlstyle{rm}\Url}\fi

\bibitem[Bhatia et~al.(2020)Bhatia, Jain, and Hooi]{exgan}
Bhatia, S., Jain, A., and Hooi, B.
\newblock Exgan: Adversarial generation of extreme samples, 2020.

\bibitem[Funk et~al.(2015)Funk, Peterson, Landsfeld, Pedreros, Verdin, Shukla,
  Husak, Rowland, Harrison, Hoell, et~al.]{funk2015climate}
Funk, C., Peterson, P., Landsfeld, M., Pedreros, D., Verdin, J., Shukla, S.,
  Husak, G., Rowland, J., Harrison, L., Hoell, A., et~al.
\newblock The climate hazards infrared precipitation with stations—a new
  environmental record for monitoring extremes.
\newblock \emph{Scientific data}, 2\penalty0 (1):\penalty0 1--21, 2015.

\bibitem[Goodfellow et~al.(2014)Goodfellow, Pouget-Abadie, Mirza, Xu,
  Warde-Farley, Ozair, Courville, and Bengio]{goodfellow}
Goodfellow, I., Pouget-Abadie, J., Mirza, M., Xu, B., Warde-Farley, D., Ozair,
  S., Courville, A., and Bengio, Y.
\newblock Generative adversarial nets.
\newblock In Ghahramani, Z., Welling, M., Cortes, C., Lawrence, N., and
  Weinberger, K.~Q. (eds.), \emph{Advances in Neural Information Processing
  Systems}, volume~27. Curran Associates, Inc., 2014.
\newblock URL
  \url{https://proceedings.neurips.cc/paper/2014/file/5ca3e9b122f61f8f06494c97b1afccf3-Paper.pdf}.

\bibitem[Higgins et~al.(2017)Higgins, Matthey, Pal, Burgess, Glorot, Botvinick,
  Mohamed, and Lerchner]{betavae}
Higgins, I., Matthey, L., Pal, A., Burgess, C.~P., Glorot, X., Botvinick, M.,
  Mohamed, S., and Lerchner, A.
\newblock beta-vae: Learning basic visual concepts with a constrained
  variational framework.
\newblock In \emph{ICLR}, 2017.

\bibitem[Kingma \& Welling(2014)Kingma and Welling]{vaes}
Kingma, D.~P. and Welling, M.
\newblock {Auto-Encoding Variational Bayes}.
\newblock In \emph{2nd International Conference on Learning Representations,
  {ICLR} 2014, Banff, AB, Canada, April 14-16, 2014, Conference Track
  Proceedings}, 2014.

\bibitem[Klemmer et~al.(2021)Klemmer, Saha, Kahl, Xu, and
  Zhu]{klemmer2021generative}
Klemmer, K., Saha, S., Kahl, M., Xu, T., and Zhu, X.~X.
\newblock Generative modeling of spatio-temporal weather patterns with extreme
  event conditioning, 2021.

\bibitem[of~Sciences~Engineering \& Medicine(2016)of~Sciences~Engineering and
  Medicine]{NAS2016}
of~Sciences~Engineering, N.~A. and Medicine.
\newblock \emph{Attribution of Extreme Weather Events in the Context of Climate
  Change}.
\newblock The National Academies Press, Washington, DC, 2016.
\newblock ISBN 978-0-309-38094-2.
\newblock \doi{10.17226/21852}.

\bibitem[Rezende et~al.(2014)Rezende, Mohamed, and
  Wierstra]{pmlr-v32-rezende14}
Rezende, D.~J., Mohamed, S., and Wierstra, D.
\newblock Stochastic backpropagation and approximate inference in deep
  generative models.
\newblock In Xing, E.~P. and Jebara, T. (eds.), \emph{Proceedings of the 31st
  International Conference on Machine Learning}, volume~32 of \emph{Proceedings
  of Machine Learning Research}, pp.\  1278--1286, Bejing, China, 22--24 Jun
  2014. PMLR.
\newblock URL \url{http://proceedings.mlr.press/v32/rezende14.html}.

\bibitem[Verdin et~al.(2018)Verdin, Rajagopalan, Kleiber, Podestá, and
  Bert]{Verdin2018}
Verdin, A., Rajagopalan, B., Kleiber, W., Podestá, G., and Bert, F.
\newblock A conditional stochastic weather generator for seasonal to
  multi-decadal simulations.
\newblock \emph{Journal of Hydrology}, 556:\penalty0 835 -- 846, 2018.
\newblock ISSN 0022-1694.

\bibitem[Wang et~al.(2021)Wang, Tang, and Gentine]{precipgan}
Wang, C., Tang, G., and Gentine, P.
\newblock Precipgan: Merging microwave and infrared data for satellite
  precipitation estimation using generative adversarial network.
\newblock \emph{Geophysical Research Letters}, 48\penalty0 (5):\penalty0
  e2020GL092032, 2021.
\newblock \doi{https://doi.org/10.1029/2020GL092032}.
\newblock URL
  \url{https://agupubs.onlinelibrary.wiley.com/doi/abs/10.1029/2020GL092032}.
\newblock e2020GL092032 2020GL092032.

\bibitem[Zadrozny et~al.(2021)Zadrozny, Watson, Szwarcman, Civitarese,
  Oliveira, Rodrigues, and Guevara]{ibm_wg}
Zadrozny, B., Watson, C.~D., Szwarcman, D., Civitarese, D., Oliveira, D.,
  Rodrigues, E., and Guevara, J.
\newblock A modular framework for extreme weather generation, 2021.

\end{thebibliography}
\bibliographystyle{icml2021}

\end{document}